\def\year{2020}\relax
\documentclass[letterpaper]{article} %
\usepackage{aaai20}  %
\usepackage{times}  %
\usepackage{helvet} %
\usepackage{courier}  %
\usepackage[hyphens]{url}  %
\usepackage{graphicx} %
\usepackage{graphicx}  %
\nocopyright
\usepackage{epsfig}
\usepackage{amsmath}
\usepackage{amssymb}
\usepackage{float}

\usepackage[pagebackref=true,breaklinks=true,letterpaper=true,colorlinks,bookmarks=false]{hyperref}

\usepackage{booktabs}
\usepackage{multirow}
\usepackage{mmstyle}
\usepackage{caption}
\usepackage{subfig}

\newlength\savewidth
\newcommand{\tablestyle}[2]{\setlength{\tabcolsep}{#1}\renewcommand{\arraystretch}{#2}\centering\footnotesize}

\title{Prime Sample Attention in Object Detection}
\author{Yuhang Cao$^{1}$ \quad Kai Chen$^{1}$ \quad Chen Change Loy$^2$ \quad Dahua Lin$^1$\\
$^1$CUHK - SenseTime Joint Lab, The Chinese University of Hong Kong \\
$^2$Nanyang Technological University\\
{{\tt\small yuhangcao@cuhk.edu.hk}\hspace{10pt}
\tt\small \{ck015,dhlin\}@ie.cuhk.edu.hk}\hspace{10pt}
{\tt\small ccloy@ntu.edu.sg})}
\begin{document}

\maketitle

\begin{abstract}
It is a common paradigm in object detection frameworks to
treat all samples equally and target at maximizing the
performance on average.
In this work, we revisit this paradigm through a careful study on
how different samples contribute to the overall performance measured
in terms of mAP. Our study suggests that the samples in each mini-batch
are neither independent nor equally important, and therefore a better
classifier on average does not necessarily mean higher mAP.
Motivated by this study, we propose the notion of Prime Samples,
those that play a key role in driving the detection performance.
We further develop a simple yet effective sampling and learning strategy
called PrIme Sample Attention (PISA) that directs the focus of
the training process towards such samples.
Our experiments demonstrate that it is often more effective to focus on
prime samples than hard samples when training a detector.
Particularly, On the MSCOCO dataset, PISA outperforms the random sampling
baseline and hard mining schemes, \eg~OHEM and Focal Loss,
consistently by around 2\% on both single-stage and two-stage detectors,
even with a strong backbone ResNeXt-101.
\end{abstract}

\section{Introduction}
\label{sec:intro}

Modern object detection frameworks,
including both single-stage~\cite{liu2016ssd,lin2017focal} and two-stage~\cite{girshick2014rich,girshick2015fast,ren2015faster},
usually adopt a region-based approach, where
a detector is trained to classify and localize sampled regions.
Therefore, the choice of region samples is critical to the success
of an object detector.
In practice, most of the samples are located in the background areas.
Hence, simply feeding all the region samples, or a random subset thereof,
through a network and optimizing the average loss is obviously not
a very effective strategy.

Recent studies~\cite{liu2016ssd,shrivastava2016training,lin2017focal} showed that focusing on difficult samples is
an effective way to boost the performance of an object detector.
A number of methods have been developed to implement this idea in
various ways.
Representative methods along this line include OHEM~\cite{shrivastava2016training}
and Focal Loss~\cite{lin2017focal}. The former explicitly selects \emph{hard samples},
\ie~those with high loss values; while the latter uses a reshaped
loss function to reweight the samples, emphasizing difficult ones.

\begin{figure}
    \centering
    \includegraphics[width=\linewidth]{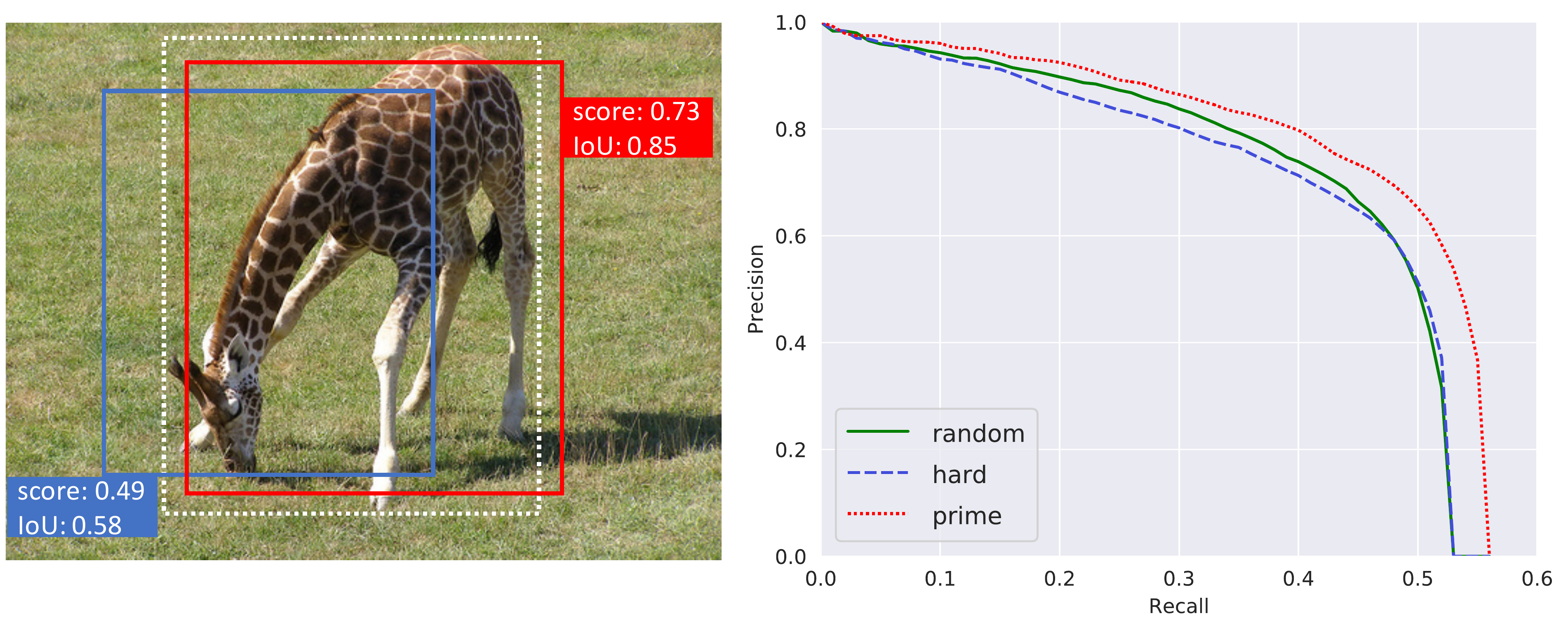}
    \caption{\small
    \textbf{Left} shows both a \emph{prime sample} (in red color) and
    a \emph{hard sample} (in blue color) for an object
    against the ground-truth. The prime sample has a high IoU
    with the ground-truth and is located more precisely around
    the object.
    \textbf{Right} shows the RoC curves obtained with different
    sampling strategies, which suggests that attending to prime samples
    instead of hard samples is a more effective strategy to boost the performance
    of a detector.}
    \label{fig:teaser}
    \vspace{-0.3cm}
\end{figure}

Though simple and widely adopted, random sampling or hard mining are not
necessarily the optimal sampling strategy in terms of training an effective
detector. Particularly, a question remains open --
\emph{what are the most important samples for training an object detector}.
In this work, we carry out a study on this issue with an aim to find
a more effective way to sample/weight regions.

Our study reveals two significant aspects that need to be taken into
consideration when designing a sampling strategy:
(1) \emph{Samples should not be treated as independent and equally important.}
Region-based object detection is to select a small subset of bounding boxes
out of a large number of candidates to cover all objects in an image.
Hence, the decisions on different samples
are competing with each other, instead of being independent (like in a classification
task). In general, it is more advisable for a detector to yield high scores
on one bounding box around each object while ensuring all objects of interest
are sufficiently covered, instead of trying to produce high scores for all
positive samples, \ie~those that substantially overlap with objects.
Particularly, our study shows that focusing on those positive samples with highest
IoUs with the ground-truth objects is an effective way towards this goal.
(2) \emph{The objective of classification and localization are correlated.}
The observation that those samples that are precisely located around
ground-truth objects are particularly important has a strong implication,
that is, the objective of classification is closely related to that
of localization. In particular, well located samples need to be
well classified with high confidences.

Inspired by the study, we propose \emph{PrIme Sample Attention (PISA)},
a simple yet effective method to sample regions and learn object detectors,
where we refer to those samples that play a more important role in
achieving high detection performance as the \emph{prime samples}.
We define \emph{Hierarchical Local Rank (HLR)} as a metric of importance.
Specifically, we use IoU-HLR to rank positive samples and Score-HLR to rank
negative samples in each mini-batch.
This ranking strategy places the positive samples with highest IoUs around each
object and the negative samples with highest scores in each cluster
to the top of the ranked list and directs the focus of the training process
to them via a simple re-weighting scheme.
We also devise a classification-aware regression loss to jointly optimize
the classification and regression branches. Particularly, this loss
would suppress those samples with large regression loss, thus reinforcing the
attention to prime samples.

We tested PISA with both two-stage and single-stage detection frameworks.
On the MSCOCO~\cite{lin2014microsoft} test-dev, with a strong backbone of
ResNet-101-32x4d, PISA improves Faster R-CNN~\cite{ren2015faster}, Mask R-CNN~\cite{he2017mask}
and RetinaNet~\cite{lin2017focal} by 2.0\%, 1.5\%, 1.8\% respectively.
For SSD, PISA achieves a gain of 2.1\%.

Our main contributions mainly lie in three aspects:
(1) Our study leads to a new insight into what samples are important for
training an object detector, thus establishing the notion of \emph{prime samples}.
(2) We devise \emph{Hierarchical Local Rank (HLR)} to rank the importance of
samples, and on top of that an importance-based reweighting scheme.
(3) We introduce a new loss called \emph{classification-aware regression loss}
that jointly optimizes both the classification and regression branches, which
further reinforces the attention to prime samples.

\section{Related Work}
\label{sec:related}

\noindent
\textbf{Region-based object detectors.}
Region-based object detectors transform the detection task into a bounding box
classification and regression problem.
Contemporary approaches mostly fall into two categories, \ie, the two-stage
and single-stage detection paradigm.
Two-stage detectors such as R-CNN~\cite{girshick2014rich}, Fast R-CNN~\cite{girshick2015fast}
and Faster R-CNN~\cite{ren2015faster} first generate a set of candidate
proposals, and then randomly sample a relatively small batch of proposals from
all the candidates. These proposals are classified into foreground classes or
background, and their locations are refined by regression.
There are also some recent improvements~\cite{dai2016r,lin2017feature,he2017mask,hu2018relation,cai2018cascade,chen2019hybrid} along this paradigm.
In contrast, single-stage detectors like SSD~\cite{liu2016ssd} and
RetinaNet~\cite{lin2017focal} directly predict class scores and box offsets
from anchors, without the region proposal step.
Other variants include~\cite{zhang2018single,li2019gradient,zhao2019m2det,zhu2019scratchdet}.
The proposed PISA is not designed for any specific detectors but can be easily
applied to both paradigms.

\noindent
\textbf{Sampling strategies in object detection.}
The most widely adopted sampling scheme in object detection is the random
sampling, that is, to randomly select some samples from all candidates.
Since negative samples are usually much more than positive ones, a fixed
ratio may be set for positive and negative samples during the sampling,
like in~\cite{girshick2015fast,ren2015faster}.
Another popular idea is to sample hard samples which have larger losses,
this strategy can lead to better optimization for classifiers.
The principle of hard mining is proposed in early detection
work~\cite{sung1998example,felzenszwalb2010object}, and also adopted by more
recent methods~\cite{liu2016ssd,girshick2014rich,shrivastava2016training} in the deep learning era.
Libra R-CNN~\cite{Libra2019CVPR} proposes IoU-balanced Sampling as an
approximation of hard negative mining.
As a special case of hard mining, DCR~\cite{cheng2018revisiting} samples hard false positive from the base classifier.
RetinaNet~\cite{lin2017focal}, does not perform actual sampling although, can
be seen as a soft version of sampling. It applies different loss weights to
samples by Focal Loss, to focus more on hard samples.
However, the goal of hard mining is to boost the average performance of
classifier and does not investigate the difference between detection and
classification. Different from that, PISA can achieve a biased performance on
different samples.
According to our study in Sec.~\ref{sec:definition}, we find that prime samples
are not necessarily hard ones, which is opposite to hard mining.

\noindent
\textbf{Improvement of NMS with localization confidence}
IoU-Net~\cite{jiang2018acquisition} proposes to use the localization confidence instead of classification scores for NMS.
It adds an extra branch to predict the IoU of samples, and use the localization
confidence, \ie, predicted IoU, for NMS.
There are some major differences between IoU-Net and our method.
Firstly, IoU-Net aims to yield higher scores for proposals with higher predicted IoU. 
In this work, we found that \emph{high IoU does not necessarily mean 
being important for training}. Particularly, the relative ranking among
proposals around the objects also plays a crucial role.
Secondly, our goal is not to improve the NMS and we do not exploit an
additional branch to predict the localization confidence, but investigate the
sample importance and propose to pay more attention to prime samples with the
importance-based reweighting, as well as a new loss to correlate the training of two branches.

\begin{figure}[t]
	\centering
	\includegraphics[width=0.9\linewidth]{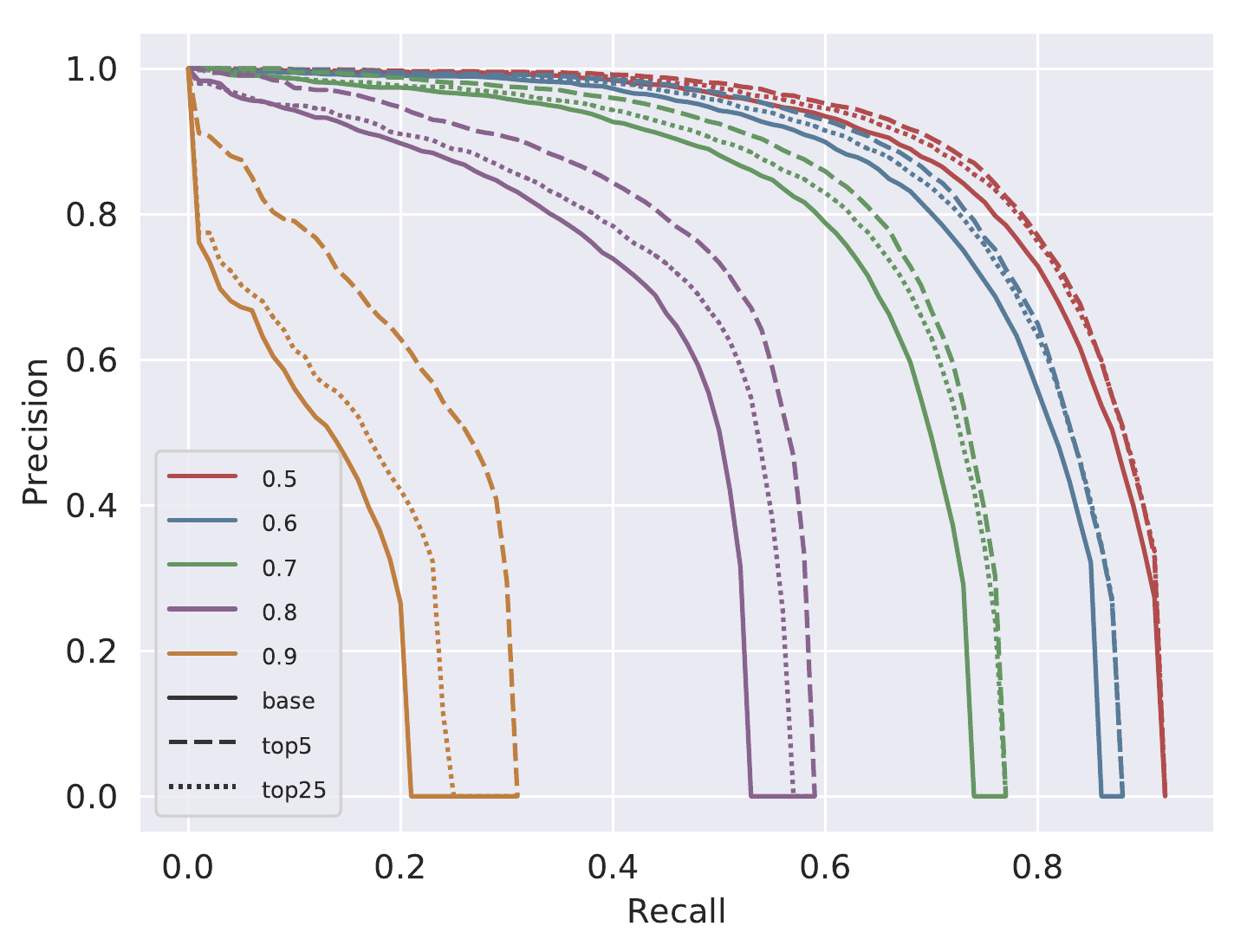}
\caption{\small Precision-recall curve under different IoU thresholds.
The solid lines correspond to the baseline, dashed lines and dotted lines
are results of reducing the classification loss by increasing scores of
positive samples. Top5 and top25 IoU-HLR samples are concentrated on
respectively.}
	\label{fig:ap_curve}
	\vspace{-0.5cm}
\end{figure}

\section{Prime Samples}
\label{sec:definition}

In this section, we introduce the concept of \emph{Prime Samples},
namely those that have greater influence on the performance of
object detection.
Specifically, we carry out a study on the importance of different
samples by revisiting how they affect mAP, the major performance metric for object detection.
Our study shows that the importance of each sample depends on how its IoU or score compares to that of the others overlapping with the same object.
Therefore, we propose HLR (IoU-HLR and Score-HLR), a new ranking strategy, as a quantitative way
to assess the importance.

\vspace{-11pt}
\paragraph{A Revisit to mAP.}
mAP is a widely adopted metric for assessing
the performance in object detection, which is computed as follows.
Given an image with annotated ground-truths, each bounding box will be marked
as true positive (TP) when:
(i) the IoU between this bounding box and its nearest ground truth is greater
than a threshold $\theta$, and
(ii) there are no other boxes with higher scores which is also a TP of the same
ground truth.
All other bounding boxes are considered as false positives (FP).
Then, the \emph{recall} is defined as the fraction of ground-truths that are
cover by TPs, and
the \emph{precision} is defined as the fraction of resulted bounding boxes
that are TPs.
On a testing dataset, one can obtain a precision-recall curve by varying the
threshold $\theta$, usually ranging from $0.5$ to $0.95$, and compute the
\emph{average precision (AP)} for each class as the area under the curve.
Then \emph{mAP} is defined as the mean of the AP values over all classes.

The way that mAP works reveals two criteria on
which positive samples are more important for an object detector.
(1) Among all bounding boxes that overlap with a ground-truth object,
the one with the highest IoU is the most important as its IoU value directly
influences the recall.
(2) Across all highest-IoU bounding boxes for different objects, the ones
with higher IoUs are more important, as they are the last ones that fall below
the IoU threshold $\theta$ as $\theta$ increases and thus have great impact
on the overall precision.

\begin{figure*}[t]
	\centering
	\includegraphics[width=\linewidth]{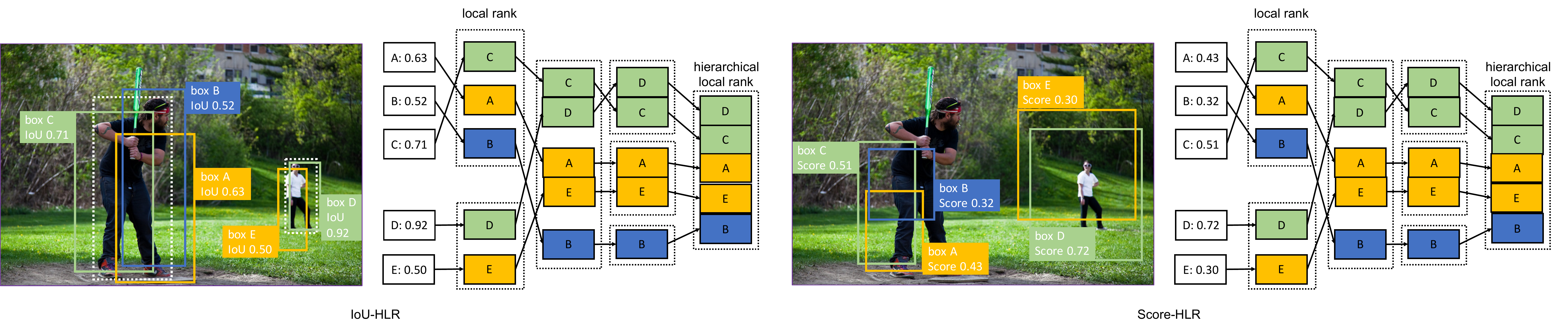}
	\caption{\small
	Two steps to compute HLR. Samples are first sorted by IoU(Score) locally,
	and then sorted in the same-rank group.}
	\label{fig:HLR}
  \vspace{-0.4cm}
\end{figure*}

\vspace{-11pt}
\paragraph{A Revisit to False Positives.}
One of the main sources of false positives are negative samples that are wrongly classified
as positive, and they are harmful to the precision and will decrease the mAP.
However, not all misclassified samples have direct influence on the final results.
During the inference, if there are multiple negative samples that heavily
overlap with each other, only the one with the highest score remains while others
are discarded after Non-Maximum Suppression (NMS).
In this way, if a negative sample is close to another one with higher score, it
becomes less important even if the score of itself may also be high because
it will not be kept in the final results.
We can learn which negative samples are important.
(1) among all negative samples within a local region, the one with the highest score is the most important.
(2) Across all highest-score samples in different regions, the ones with
higher scores are more important, because they are the first ones that decrease
the precision.

\vspace{-11pt}
\paragraph{Hierarchical Local Rank (HLR).}
Based on the analysis above, we propose \emph{IoU Hierarchical Local Rank (IoU-HLR)} and \emph{Score Hierarchical Local Rank (Score-HLR)}
to rank the importance of positive and negative samples in a mini-batch.
This rank is computed in a hierarchical manner, which reflects the
relation both locally (around each ground truth object or some local region)
and globally (over the whole image or mini-batch).
Notably, We compute IoU-HLR and Score-HLR based on the final located position
of samples, other than the bounding box coordinates before regression,
since mAP is evaluated based on the regressed samples.

As shown in Figure~\ref{fig:HLR}, to compute IoU-HLR, we first divide all samples into different groups, based on their nearest ground truth objects.
Next, we sort the samples within each group in descending order by their IoU
with the ground truth, and get the IoU Local Rank (IoU-LR).
Subsequently, we take samples with the same IoU-LR and sort them in descending
order. Specifically, all top-1 IoU-LR samples are collected and sorted, followed
by top2, top3, and so on.
These two steps result in the ranking among all samples, that is the \emph{IoU-HLR}.
IoU-HLR follows the two criteria mentioned above.
First, it places the positive samples with higher local ranks ahead, which are the
samples that are most important to each individual ground-truth object.
Second, within each local group, it re-ranks the samples according to
IoU, which aligns with the second criterion.
Note that it is often good enough to ensure high accuracies on those samples
that top this ranked list as they directly influence both the recall and the
precision, especially when the IoU threshold is high.
As shown in Figure~\ref{fig:ap_curve}, the solid lines are the precision-recall
curves under different IoU thresholds.
We simulate some experiments by increasing the scores of samples.
With the same budget, \eg, reducing the total loss by 10\%, we increase the scores of top5 and top25 IoU-HLR samples and plot the results in
dashed and dotted lines respectively.
The results suggest that focusing on only top samples is better than
attending more samples equally.

\begin{figure}[t]
	\centering
	\includegraphics[width=\linewidth]{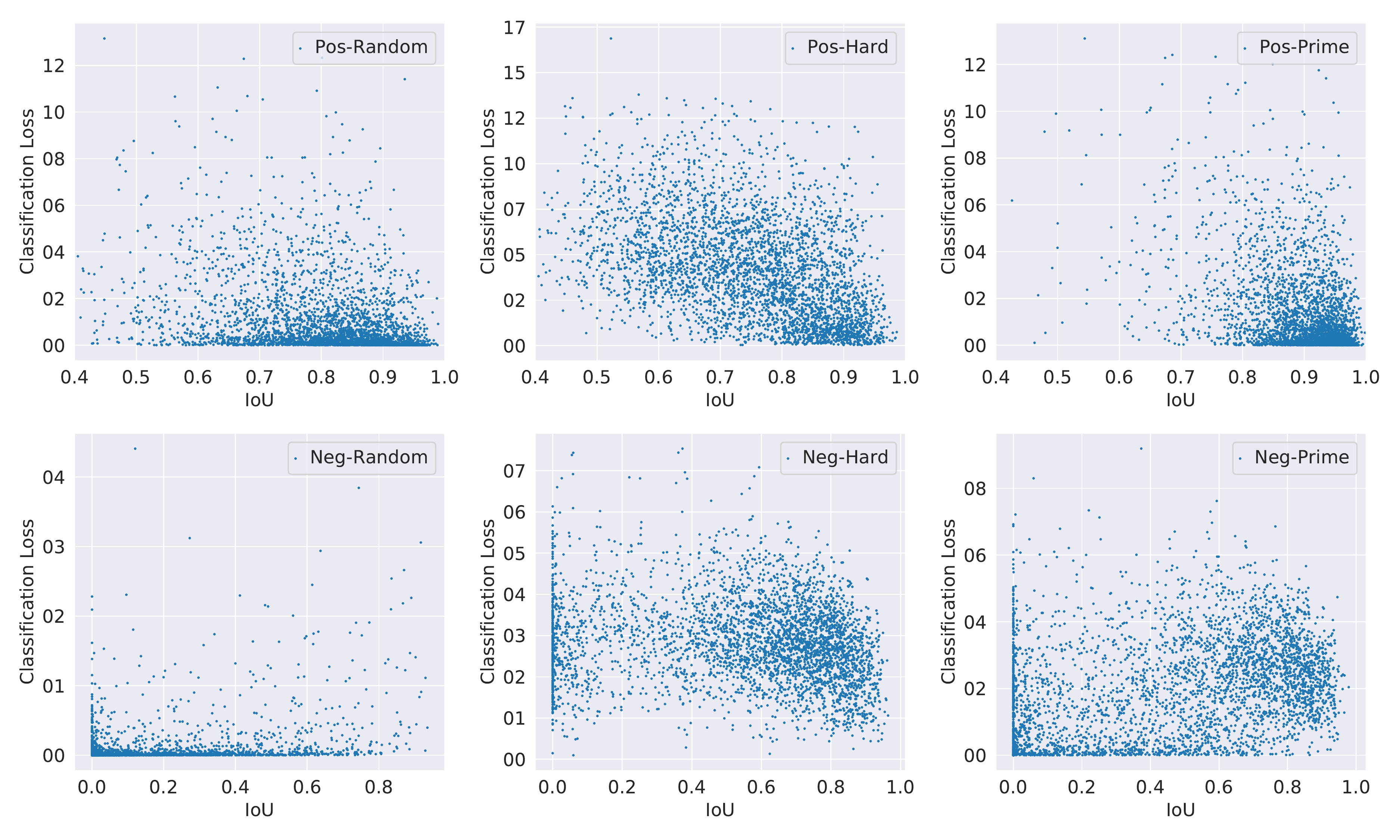}
	\caption{\small The distribution of random, hard, and prime samples.
	Here the hard samples are the ones with top-3 loss values
	from each image; while the prime samples are those ranked as top-3
	HLRs.}
	\label{fig:sample-scatter}
  \vspace{-0.4cm}
\end{figure}

We compute Score-HLR for negative samples in a similar way to IoU-HLR.
Unlike positive samples that are naturally grouped by each ground truth object,
negative samples may also appear on background regions,
thus we first group them into different clusters with NMS.
We adopt the maximum score over all foreground classes as the score of
negative samples and then follow the same steps as computing IoU-HLR,
as shown in Figure~\ref{fig:HLR}.

We plot the distributions of random, hard, and prime samples in
Figure~\ref{fig:sample-scatter}, with the IoU \textit{vs.} classification loss.
The top row shows positive samples and the bottom row presents negative ones.
It is observed that hard positive samples tend to have high classification
losses and scatter over a wider range along the IoU axis, while prime positive
samples tend to have high IoUs and low classification losses.
Hard negative samples tend to have high classification losses and high IoUs,
while prime negative samples also cover some low loss samples and have a more diverged IoU distribution.
This suggests that these two categories of samples are of
essentially different characteristics.

\begin{figure*}[ht]
	\centering
	\includegraphics[width=0.9\linewidth]{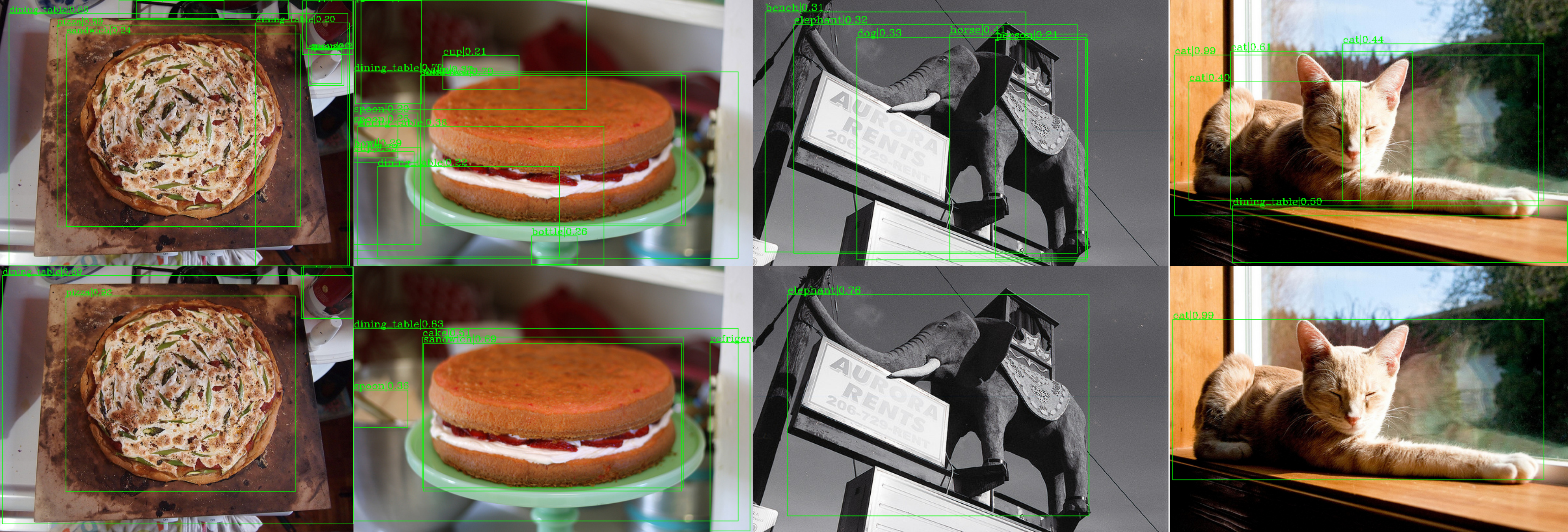}
    \caption{
        Examples of PISA (bottom) and random sampling (top) results.
        The score threshold for visualization is 0.2.}
	\label{fig:examples}
	\vspace{-0.4cm}
\end{figure*}

\section{Learn Detectors via Prime Sample Attention}
\label{sec:methodology}

The aim of object detection is not to obtain a better classification accuracy
on average, but to achieve as good performance as possible on prime samples in
the set, as discussed in Section~\ref{sec:definition}.
In this work, we propose Prime Sample Attention, a simple and effective
sampling and learning strategy which pay more attention to prime samples.
PISA consists of two components: Importance-based Sample Reweighting (ISR) and
Classification Aware Regression Loss (CARL).
With the proposed method, the training process is biased on prime samples other
than evenly treat all ones.
Firstly, the loss weight of prime samples are larger than the others,
so that the classifier tends to be more accurate on these samples.
Secondly, the classifier and regressor are learned with a joint objective,
thus scores of positive prime samples get boosted relative to unimportant ones.

\subsection{Importance-based Sample Reweighting}
\label{subsec:ISR}

Given the same classifier, the distribution of performance usually matches the
distribution of training samples. If part of the samples occurs more frequently
in the training data, a better classification accuracy on those samples is
supposed to be achieved.
Hard sampling and soft sampling are two different ways to change the training
data distribution. Hard sampling selects a subset of samples from all
candidates to train a model, while soft sampling assigns different weights for
all samples. Hard sampling can be seen as a special case of soft sampling,
where each sample is assigned a loss weight of either 0 or 1.

To make fewer modifications and fit existing frameworks, we propose a soft
sampling strategy named Importance-based Sample Reweighting (ISR), which assigns
different loss weights to samples according to their importance.
ISR consists of positive sample reweighting and negative sample reweighting, denoted as ISR-P and ISR-N, respectively.
We adopt IoU-HLR as the importance measurement for positive samples, and Score-HLR for negative samples.
Given the importance measurement, the remaining
question is how to map the importance to an appropriate loss weight.

We first transform the rank to a real value with a linear mapping.
According to its definition, HLR is computed separately within each class (N foreground classes and 1 background class).
For class $j$, supposing there are $n_j$ samples in total with the HLR
$\{r_1, r_2, \dots, r_{n_j}\}$, where $0\leq r_i\leq n_j-1$,
we use a linear function to transform each $r_i$ to $u_i$ as shown in Equ.~\ref{equ:linear-mapping}.
Here $u_i$ denotes the importance value of the $i$-th sample of class $j$.
$n_{max}$ denotes the max value of $n_j$ over all classes, 
which ensures the samples at the same rank of different classes will be assigned the same $u_i$.

\begin{equation}
    \label{equ:linear-mapping}
    u_i = \frac{n_{max} - r_i}{n_{max}}
\end{equation}
A monotone increasing function is needed to further cast the sample importance
$u_i$ to a loss weight $w_i$. Here we adopt an exponential form as
Equ.~\ref{equ:exp-isr}, where $\gamma$ is the degree factor indicating how much
preference will be given to important samples and $\beta$ is a bias that
decides the minimum sample weight.
\begin{equation}
    \label{equ:exp-isr}
    w_i = ((1-\beta)u_i + \beta)^\gamma
\end{equation}

With the proposed reweighting scheme, the cross entropy classification loss
can be rewritten as Equ.~\ref{equ:cls-loss}, where $n$ and $m$ are the
numbers of positive and negative samples respectively, $s$ and $\hat{s}$
denote the predicted score and classification target.
Note that simply adding loss weights will change the total value of losses and
the ratio between the loss of positive and negative samples, so we normalize
$w$ to $w^\prime$ in order to keep the total loss unchanged.
\begin{equation}
    \label{equ:cls-loss}
    \begin{split}
        L_{cls} &= \sum_{i=1}^n w_i^\prime CE(s_i, \hat{s_i}) +
                    \sum_{j=1}^{m} w_j^\prime CE(s_j, \hat{s_j}) \\
        w_i^\prime &= w_i\frac{\sum_{i=1}^n CE(s_i, \hat{s_i})}{\sum_{i=1}^n w_i CE(s_i, \hat{s_i})} \\
        w_j^\prime &= w_j\frac{\sum_{j=1}^m CE(s_j, \hat{s_j})}{\sum_{i=j}^m w_j CE(s_j, \hat{s_j})}
    \end{split}
\end{equation}

\subsection{Classification-Aware Regression Loss}
\label{subsec:CARL}

Re-weighting the classification loss is a straightforward way to focus on prime samples.
Besides that, we develop another method to highlight the prime samples,
motivated by the earlier discussion that classification and localization is correlated.
We propose to jointly optimize the two branches with a Classification-Aware
Regression Loss (CARL).
CARL can boost the scores of prime samples while suppressing the scores of other ones.
The regression quality determines the importance of a sample and we
expect the classifier to output higher scores for important samples.
The optimization of two branches should be correlated other than independent.

Our solution is to add a classification-aware regression loss,
so that gradients are propagated from the regression branch to the classification branch.
To this end, we propose CARL as shown in Equ.~\ref{equ:reg-loss}.
$p_i$ denotes the predicted probability of the corresponding ground truth class
and $d_i$ denotes the output regression offset.
We use an exponential function to transform the $p_i$ to $v_i$, and then
rescale it according to the average value of all samples.
$\cL$ is the commonly used smooth L1 loss.

\begin{equation}
    \label{equ:reg-loss}
    \begin{split}
        L_{carl} &= \sum_{i=1}^n c_i \cL(d_i, \hat{d_i}) \\
        c_i &= \frac{v_i}{\frac{1}{n}\sum_{i=1}^n v_i} \\
        v_i &= ((1-b)p_i + b)^k
    \end{split}
\end{equation}

It is obvious that the gradient of $c_i$ is proportional to the original
regression loss $\cL(d_i, \hat{d_i})$.
In the supplementary, we prove that there is a positive correlation between $\cL(d_i, \hat{d_i})$ and the gradient of $p_i$.
Namely, samples with greater regression loss will receive large gradients
for the classification scores, which means stronger suppression on the
classification scores.
In another view, $\cL(d_i, \hat{d_i})$ reflects the localization quality of
sample $i$, thus can be seen as an estimation of IoU and further seen as an
estimation of IoU-HLR.
Approximately, top ranked samples have low regression loss, thus the gradients
of classification scores are smaller.
With CARL, the classification branch gets supervised by the regression loss.
The scores of unimportant samples are greatly suppressed, while the attention
to prime samples are reinforced.

\section{Experiments}
\label{sec:experiments}

\begin{table*}[htb]
	\centering
	\tablestyle{6pt}{1.05}
	\caption{Results of different detectors on COCO \emph{test-dev}.}
	\label{tab:results}
	\begin{tabular}{l|l| c | c c c c c c}
		\hline
		Method & Backbone & AP & $\text{AP}_{50}$ & $\text{AP}_{75}$ & $\text{AP}_{S}$ & $\text{AP}_{M}$ & $\text{AP}_{L}$ \\
		\hline
		\emph{Two-stage detectors} & & & & & & & &\\
		Faster R-CNN          & ResNet-50   & 36.7 & 58.8 & 39.6 & 21.6 & 39.8 & 44.9 \\
		Faster R-CNN          & ResNeXt-101 & 40.3 & 62.7 & 44.0 & 24.4 & 43.7 & 49.8 \\
		Mask R-CNN            & ResNet-50   & 37.5 & 59.4 & 40.7 & 22.1 & 40.6 & 46.2 \\
		Mask R-CNN            & ResNeXt-101 & 41.4 & 63.4 & 45.2 & 24.5 & 44.9 & 51.8 \\
		\hline
		Faster R-CNN w/ PISA  & ResNet-50   & \textbf{38.8}(+2.1) & 59.3 & 42.7 & 22.1 & 41.7 & 48.8 \\
		Faster R-CNN  w/ PISA & ResNeXt-101 & \textbf{42.3}(+2.0) & 62.9 & 46.8 & 24.8 & 45.5 & 53.1 \\
		Mask R-CNN  w/ PISA   & ResNet-50   & \textbf{39.3}(+1.8) & 59.6 & 43.5 & 22.1 & 42.3 & 49.4 \\
		Mask R-CNN  w/ PISA   & ResNeXt-101 & \textbf{42.9}(+1.5) & 63.2 & 47.4 & 24.9 & 46.2 & 54.0 \\
		\hline
		\hline
		\emph{Single-stage detectors} & & & & & & & & \\
		RetinaNet & ResNet-50          & 35.9 & 56.0 & 38.3 & 19.8 & 38.9 & 45.0 \\
		RetinaNet & ResNeXt-101        & 39.0 & 59.7 & 41.9 & 22.3 & 42.5 & 48.9 \\
		SSD300    & VGG16              & 25.7 & 44.2 & 26.4 & 7.0  & 27.1 & 41.5 \\
		SSD512    & VGG16              & 29.6 & 49.5 & 31.2 & 11.7 & 33.0 & 44.1 \\
		\hline
		RetinaNet w/ PISA & ResNet-50   & \textbf{37.3}(+1.4) & 56.5 & 40.3 & 20.3 & 40.4 & 47.2 \\
		RetinaNet w/ PISA & ResNeXt-101 & \textbf{40.8}(+1.8) & 60.5 & 44.2 & 23.0 & 44.2 & 51.4 \\
		SSD300 w/ PISA    & VGG16       & \textbf{27.7}(+2.0) & 45.3 & 29.2 & 8.3  & 29.1 & 44.1 \\
		SSD512 w/ PISA    & VGG16       & \textbf{31.7}(+2.1) & 50.5 & 33.9 & 13.0 & 35.1 & 46.1 \\
		\hline
	\end{tabular}
	\vspace{-0.2cm}
\end{table*}

\begin{table}[htb]
	\centering
	\tablestyle{6pt}{1.1}
	\caption{Results of different detectors on VOC2007 test.}
	\label{tab:voc-results}
	\begin{tabular}{l | l | c | c}
		\hline
		Method & Backbone & AP(VOC) & AP(COCO)\\
		\hline
		Faster R-CNN      & ResNet-50 & 79.1 & 48.4 \\
		Faster R-CNN w/ PISA & ResNet-50 & 81.2 & \textbf{52.3} \\
		RetinaNet         & ResNet-50 & 79.0 & 51.8 \\
		RetinaNet w/ PISA & ResNet-50 & 79.3 & \textbf{54.0} \\
		\hline
	\end{tabular}
	\vspace{-0.2cm}
\end{table}

\subsection{Experimental Setting}
\noindent
\textbf{Dataset and evaluation metric.}
We conduct experiments on the challenging MS COCO 2017 dataset~\cite{lin2014microsoft}.
It consists of two subsets: the \emph{train} split with 118k images and
\emph{val} split with 5k images.
We use the \emph{train} split for training and report the performance on
\emph{val} and \emph{test-dev}.
The standard COCO-style AP metric is adopted, which averages mAP of 
IoUs from 0.5 to 0.95 with an interval of 0.05.

\noindent
\textbf{Implementation details.}
We implement our methods based on mmdetection~\cite{mmdetection}.
ResNet-50~\cite{he2016deep}, ResNeXt-101-32x4d~\cite{xie2017aggregated}, 
VGG16~\cite{simonyan2014very} are adopted as backbones in our experiments.
Detailed settings are described in the supplementary material.

\subsection{Results}

\noindent\textbf{Overall results.}
We evaluate the proposed PISA on both two-stage and single-stage detectors,
on two popular benchmarks.
We use the same hyper-parameters of PISA for all backbones and datasets.
The results on MS COCO dataset are shown in Table~\ref{tab:results}.
PISA achieves consistent mAP improvements on all detectors with different
backbones, indicating its effectiveness and generality.
Specifically, it improves Faster R-CNN, Mask R-CNN and RetinaNet by 2.1\%, 1.8\% and 1.4\% with a ResNet-50 backbone.
Even with a strong backbone like ResNeXt-101-32x4d, similar improvements are observed.
On SSD300 and SSD512, the gain is more than 2.0\%.
Notably, PISA introduces no additional parameters and the inference time remains the same as the baseline.

On the PASCAL VOC dataset, PISA also outperforms the baselines,
as shown in Table~\ref{tab:voc-results}.
PISA not only brings performance gains under the VOC evaluation metric that use 0.5 as
the IoU threshold, but achieves significant better under the COCO metric that
use the average of multiple IoU thresholds.
This implies that PISA is especially beneficial to high IoU metrics and makes
more accurate prediction on precisely located samples.

\noindent\textbf{Comparison of different sampling methods.}
To investigate the effects of different sampling methods, 
we apply random sampling (R), hard mining (H) and PISA (P) on
positive and negative samples respectively.
Faster R-CNN is adopted as the baseline methods.
As shown in Table~\ref{tab:different-sampling}, PISA outperforms random sampling and hard mining in all cases.
For positive samples, PISA achieves 1.6\% higher mAP than random sampling and 2.0\% higher than hard mining.
For negative samples, PISA surpasses them by 0.9\% and 0.4\%, respectively.
When applying to both positive and negative samples, PISA leads to 2.1\% and 1.7\% improvements compared to random sampling and hard mining.
It is noted that the gain mainly originates from the AP of high IoU thresholds,
such as $\text{AP}_{75}$. 
This indicates that attending prime samples helps the classifier to be
more accurate on samples with high IoUs.
We demonstrate some qualitative results of PISA and the baseline in Figure~\ref{fig:examples}. PISA results in less false positives and higher scores for prime positive samples.

\begin{table}[t]
	\centering
	\caption{Comparison of different sampling strategies. Results are evaluated on COCO \emph{val}.}
    \label{tab:different-sampling}
	\tablestyle{6pt}{1.1}
	\begin{tabular}{*{2}{c}|*{6}{c}}
		\hline
		pos & neg & AP   & $\text{AP}_{50}$ & $\text{AP}_{75}$ & $\text{AP}_{S}$ & $\text{AP}_{M}$ & $\text{AP}_{L}$ \\
		\hline
		R            & R & 36.4 & 58.4 & 39.1 & 21.6 & 40.1 & 46.6 \\
		H            & R & 36.0 & 58.3 & 38.7 & 21.1 & 39.5 & 45.8 \\
		P   & R & \textbf{38.0} & \textbf{58.5} & \textbf{41.7} & \textbf{22.4} & \textbf{41.6} & \textbf{48.3} \\
		\hline
		R &        H & 36.9 & 58.2 & 40.1 & 21.2 & \textbf{40.7} & 48.5 \\
		R & P   & \textbf{37.3} & \textbf{58.8} & \textbf{40.6} & \textbf{21.7} & 40.6 & \textbf{48.7} \\
		\hline
		H            & H            & 36.8 & 58.2 & 39.8 & 21.2 & 40.4 & 48.5 \\
		P   & P   & \textbf{38.5} & \textbf{58.8} & \textbf{42.3} & \textbf{22.2} & \textbf{41.5} & \textbf{50.8} \\
		\hline
	\end{tabular}
\end{table}

\subsection{Analysis}

We perform a thorough study on each component of PISA, and explain
how it works compared with random sampling and hard mining.

\noindent\textbf{Component Analysis.}
Table~\ref{tab:components} shows the effects of each component of PISA.
We can learn that ISR-P, ISR-N and CARL improve the mAP by 0.7\%, 0.9\%, 1.0\% respectively. 
ISR (ISR-P + ISR-N) boots mAP 1.5\%. 
Applying PISA only to positive samples (ISR-P + CARL) increases mAP by 1.6\%.
With all 3 components, PISA achieves a total gain of $2.1\%$.

\begin{table} [tb]
	\centering
	\tablestyle{6pt}{1.1}
	\caption{Effectiveness of components of PISA.}
	\label{tab:components}
	\addtolength{\tabcolsep}{-2.5pt}
	\begin{tabular}{c c c| c c c c c c}
		\hline
		ISR-P      & ISR-N      & CARL       & AP   & AP$_{50}$ & AP$_{75}$ & AP$_\text{S}$ & AP$_\text{M}$ & AP$_\text{L}$ \\
		\hline
		           &            &            & 36.4 & 58.4 & 39.1 & 21.6 & 40.1 & 46.6 \\
		\checkmark &            &            & 37.1 & 58.7 & 40.3 & 21.7 & 40.9 & 47.1 \\
				   & \checkmark &            & 37.3 & 58.8 & 40.6 & 21.7 & 40.6 & 48.7 \\
				   &            & \checkmark & 37.4 & 57.9 & 41.2 & 22.1 & 41.1 & 47.7 \\
		\checkmark & \checkmark &            & 37.9 & \textbf{59.4} & 41.6 & 21.7 & 41.2 & 49.7 \\
		\checkmark &            & \checkmark & 38.0 & 58.5 & 41.7 & \textbf{22.4} & \textbf{41.6} & 48.3 \\
		\checkmark & \checkmark & \checkmark & \textbf{38.5} & 58.8 & \textbf{42.3} & 22.2 & 41.5 & \textbf{50.8} \\
    \hline
	\end{tabular}
\end{table}

\noindent\textbf{Ablation experiments of hyper-parameters.}
For both ISR and CARL, we use an exponential transformation function of
Equ.~\ref{equ:exp-isr} and 2 hyper-parameters
($\gamma_P,\beta_P$ for ISR-P, $\gamma_N,\beta_N$ for ISR-N, and $k,b$ for CARL) are introduced.
The exponential factor $\gamma$ or $k$ controls the steepness of the curve, 
while the constant factor $\beta$ or $b$ affects the minimum value.

When performing ablation study on hyper-parameters of ISR-P, ISR-N or CARL,
we do not involve other components.
A larger $\gamma$ and small $\beta$ means larger gap between prime samples and
unimportant samples, so that we are more focus on prime samples.
The opposite case means we pay more equal attention to all samples.
Through a coarse search, we adopt $\gamma_P=2.0, \gamma_N=0.5, \beta_P=\beta_N=0$ for ISR, and $k=1.0,b=0.2$ for CARL.
We also observe that the performance is not very sensitive to those hyper-parameters.

\begin{table} [tb]
	\centering
	\tablestyle{7pt}{1.1}
    \caption{Varying $\gamma,\beta$ in ISR and $k,b$ in CARL.}
    \label{tab:reweighting}
    \addtolength{\tabcolsep}{-1pt}
    \begin{tabular}{c c | c || c c | c || c c | c }
        \hline
        $\gamma_P$ & $\beta_P$ & AP & $\gamma_N$ & $\beta_N$ & AP   & $k$ & $b$ & AP \\
        \hline
        0.5 & 0.0 & 36.9          & 0.5 & 0.0 & \textbf{37.3} & 0.5 & 0.0 & 37.3 \\
        1.0 & 0.0 & 36.9          & 1.0 & 0.0 & 37.2          & 1.0 & 0.0 & 37.4 \\
        2.0 & 0.0 & \textbf{37.1} & 2.0 & 0.0 & 37.1          & 2.0 & 0.0 & N/A   \\
        \hline
        2.0 & 0.1 & 37.0          & 0.5 & 0.1 & 37.2          & 1.0 & 0.1 & 37.4 \\
        2.0 & 0.2 & 36.8          & 0.5 & 0.2 & 37.1          & 1.0 & 0.2 & \textbf{37.4} \\
        2.0 & 0.3 & 36.9          & 0.5 & 0.3 & 37.2          & 1.0 & 0.3 & 37.2 \\
        \hline
    \end{tabular}
\end{table}

\begin{figure}[t]
	\centering
	\includegraphics[width=\linewidth]{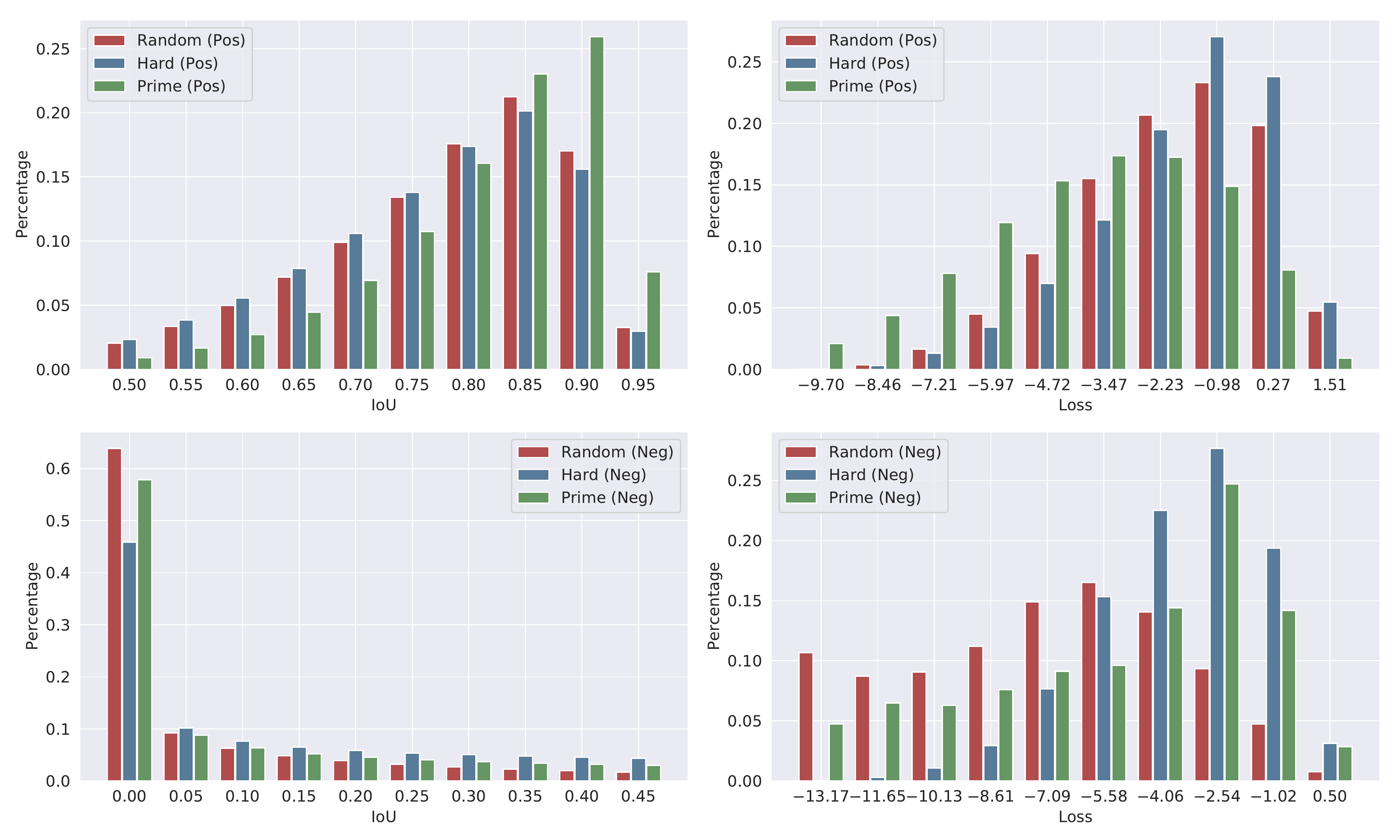}
	\caption{IoU and Loss distribution of random, hard, and prime samples.}
	\label{fig:sample_analysis}
  \vspace{-0.4cm}
\end{figure}

\noindent\textbf{What samples do different sampling strategies prefer?}
To understand how ISR works, we study the sample distribution of different sampling strategies from the aspects of IoU and loss.
Sample weights are taken into account when obtaining the distribution.
Results are shown in Figure~\ref{fig:sample_analysis}.
For positive samples, we can learn that samples selected by hard mining 
and PISA diverge from each other.
Hard samples have high losses and low IoUs, while prime samples come with
high IoUs and low losses, indicating that prime samples tend to be easier
for classifiers.
For negative samples, PISA presents an intermediate preference between random sampling and hard mining.
Unlike random sampling that focus more on low IoU and easy samples or hard mining that attend relatively high IoU and hard samples, PISA maintains the diversity of samples.

\begin{figure} [tb]
    \centering
	\includegraphics[width=\linewidth]{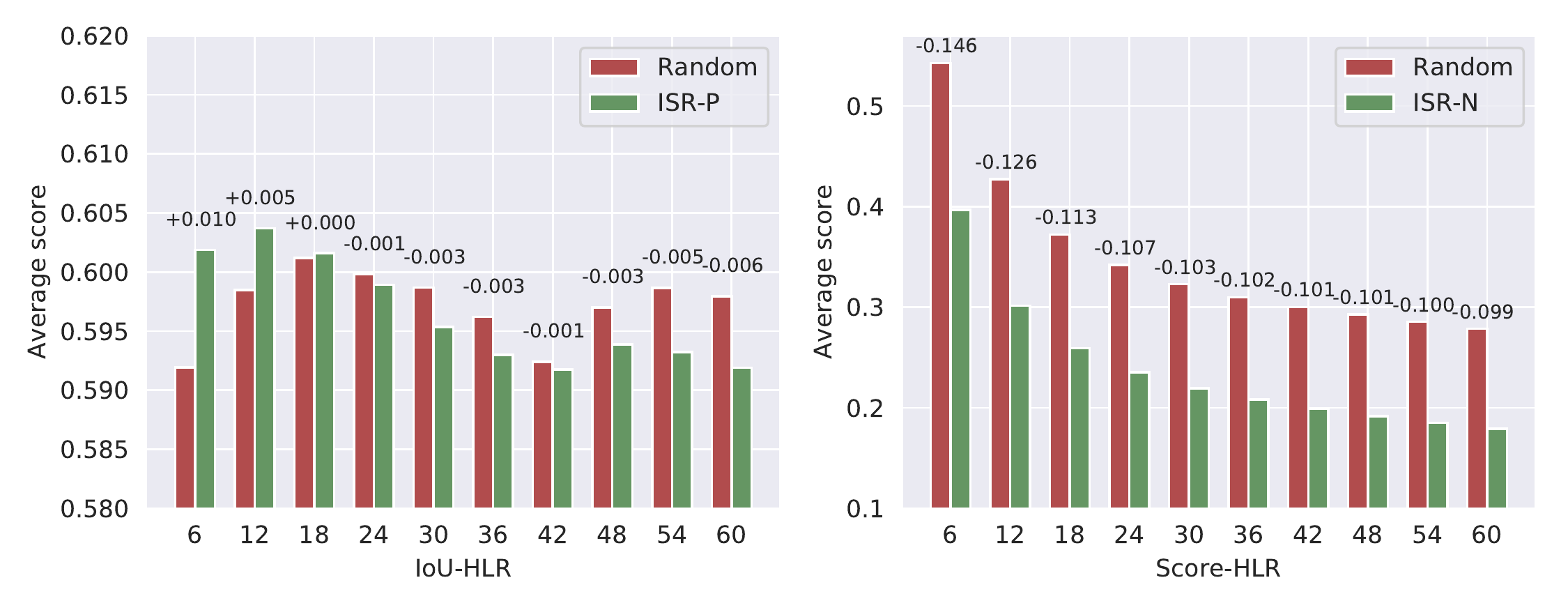}
	\caption{Affects of ISR on samples scores. 
			 Left: average scores of positive samples vary with IoU-HLR.
			 Right: average scores of negative samples vary with Score-HLR.}
	\vspace{-0.3cm}
    \label{fig:HLR-Scores}
\end{figure}

\noindent\textbf{How does ISR affect classification scores?}
ISR assigns larger weights to prime samples, but does it achieve the biased
classification performance as expected?
In Figure~\ref{fig:HLR-Scores}, we plot the score distribution of positive and
negative samples \wrt~different HLRs.
For positive samples, the scores of top-ranked samples are higher than
the baseline, while that of lower-ranked samples are lower.
The result demonstrates ISR-P biases the classifier,
thus boosting the prime samples while suppressing others.
For negative samples, the scores of all samples are lower than the baseline,
especially for top-ranked samples. This implies that ISR-N has a strong
suppression for false positives.

\begin{figure} [tb]
    \centering
    \includegraphics[width=0.9\linewidth]{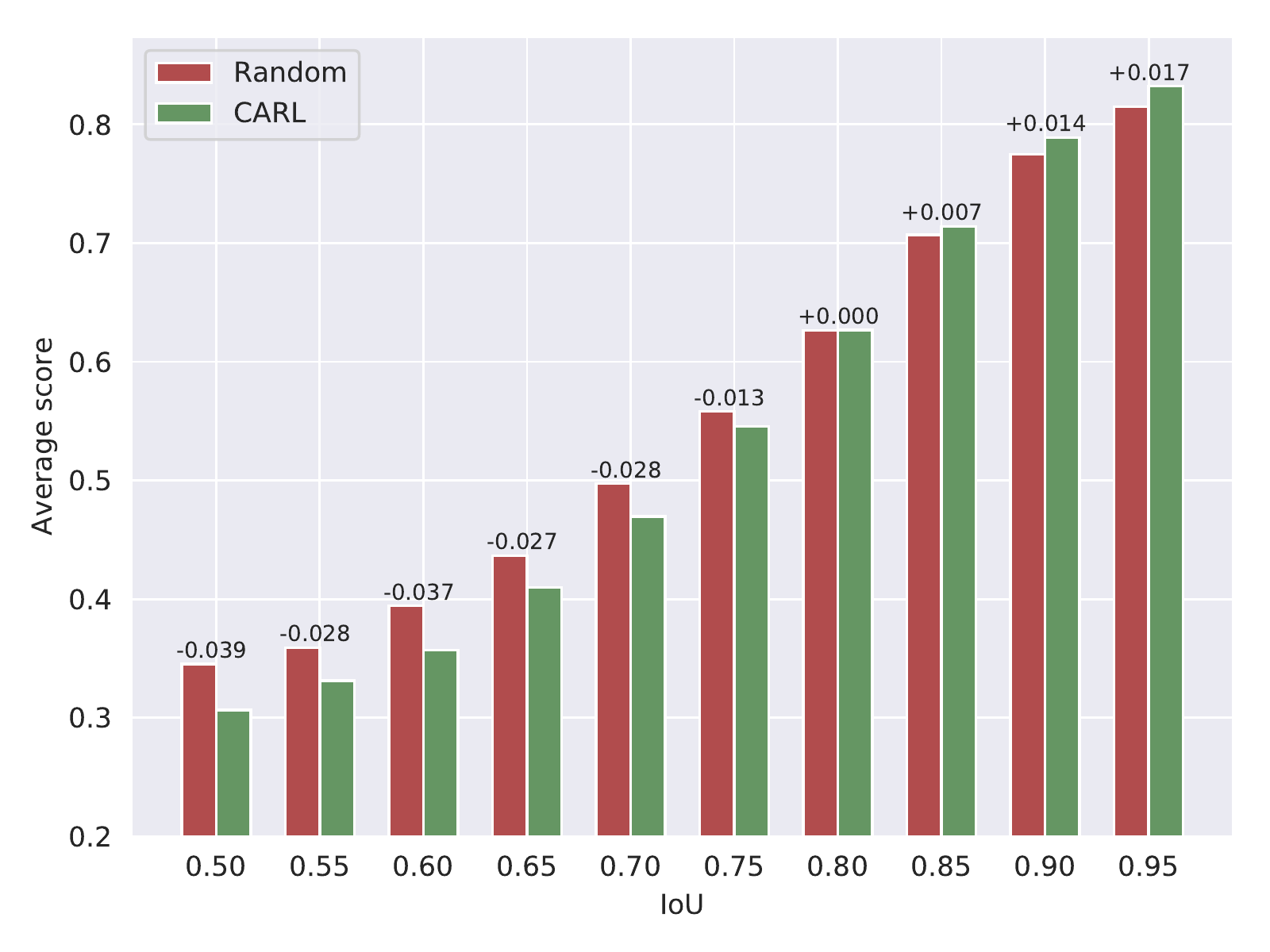}
	\caption{Affects of CARL on the scores of positive samples vary with IoU interval.}
	\vspace{-0.5cm}
    \label{fig:CARL}
\end{figure}

\noindent\textbf{How does CARL affect classification scores?}
CARL correlates the classification and localization branches by introducing the
classification scores to the regression loss.
The gradient will suppress the scores of samples with lower regression quality,
but highlight the prime samples that are localized more accurately.
Figure~\ref{fig:CARL} shows the scores of samples of different IoUs.
Compared with the FPN baseline, 
CARL boosts scores of high IoU samples but decreases scores of low IoU ones as expected.

\section{Conclusion}

We study the question what are the most important samples for training an
object detector, and establishing the notion of \emph{prime samples}.
We present PrIme Sample Attention (PISA), a simple and effective sampling and learning strategy to highlight important samples.
On both MS COCO and PASCAL VOC dataset, PISA achieves consistent improvements
over random sampling and hard mining counterparts.

\section*{Appendix A: Derivative of CARL}

As discussed in Section 4.2, we prove that there is a positive correlation
between $\frac{\partial L_{carl}}{\partial p_i}$ and $\cL(d_i, \hat{d_i})$, where

\begin{equation}
  \label{equ:reg-loss}
  \begin{split}
      L_{carl} &= \sum_{i=1}^n c_i \cL(d_i, \hat{d_i}) \\
      c_i &= \frac{v_i}{\frac{1}{n}\sum_{i=1}^n v_i} \\
      v_i &= ((1-b)p_i + b)^k
  \end{split}
\end{equation}

\noindent By chain rule,

\begin{equation}
  \begin{split}
  \frac{\partial L_{carl}}{\partial p_i}
       &= \frac{\partial \cL_{carl}}{\partial c_i}\frac{\partial c_i}{\partial p_i} \\
       &= \cL(d_i, \hat{d_i}) \frac{\partial c_i}{\partial p_i}
  \end{split}
  \label{eq:5}
\end{equation}
Furthermore,
\begin{equation}
    \frac{\partial c_i}{\partial p_i}
    = \frac{\partial c_i}{\partial v_i} \frac{\partial v_i}{\partial p_i}
    \label{eq:6}
\end{equation}
Denoting $S=\sum_{i=1}^n v_i$, we have
\begin{equation}
    \frac{\partial c_i}{\partial v_i} = \frac{n}{S} (1 - \frac{v_i}{S})
    \label{eq:7}
\end{equation}
The batch size is usually large, so $v_i << S$. Thus we have 
\begin{equation}
  \frac{\partial c_i}{\partial v_i} \approx \frac{n}{S}
  \label{eq:8}
\end{equation}
On the other hand,
\begin{equation}
  \frac{\partial v_i}{\partial p_i} = (1 - b)k((1 - b)p_i + b)^{k - 1}
  \label{eq:9}
\end{equation}
We have $0\leq b<1$ and $k>0$, so $\frac{\partial v_i}{\partial p_i} > 0$.
Especially when $k=1$, $\frac{\partial v_i}{\partial p_i} = 1-b$.

Combining \eqref{eq:5}\eqref{eq:6}\eqref{eq:8}\eqref{eq:9},
\begin{equation}
  \begin{split}
    \frac{\partial L_{carl}}{\partial p_i}
    &= \frac{n}{S}\frac{\partial v_i}{\partial p_i} \cL(d_i, \hat{d_i}) 
  \end{split}
\end{equation}
When $k=1$, $\frac{\partial L_{carl}}{\partial p_i}=\frac{n(1-b)}{S}\cL(d_i, \hat{d_i})$, indicating that $\frac{\partial L_{carl}}{\partial p_i}$ is proportional to $\cL(d_i, \hat{d_i})$, otherwise $\frac{\partial L_{carl}}{\partial p_i}$ and $\cL(d_i, \hat{d_i})$ are positively correlated.

\vspace{0.4cm}

\section*{Appendix B: Implementation details}
We use 8 Tesla V100 GPUs in all experiments.
For SSD, we train the model for a total of 120 epochs with a minibatch of 64 images (8 images per GPU). 
The learning rate is initialized as 0.001 and decreased by 0.1 after 80 and 110 epochs.
For other methods, we adopt ResNet-50 or ResNeXt-101-32x4d as the backbone. FPN is used by default.
The batch size is 16 (2 images per GPU). 
Models are trained for 12 epochs with an initial learning rate of 0.02, 
which is decreased by 0.1 after 8 and 11 epochs respectively.
We sample 512 RoIs from 2000
proposals and the ratio of positive/negative samples is set to 1:3.
PISA consists of ISR (ISR-P and ISR-N) and CARL with one exception, \ie,
ISR-N is not used in single-stage models
because the number of negative samples in single-stage models are much greater than two-stage ones, which will
intruduce significant overhead for training time.

{\small
    \bibliographystyle{aaai}
	\bibliography{sections/egbib}
}

\end{document}